# OCR POST-PROCESSING ERROR CORRECTION ALGORITHM USING GOOGLE'S ONLINE SPELLING SUGGESTION

Youssef Bassil, Mohammad Alwani
LACSC – Lebanese Association for Computational Sciences
Registered under No. 957, 2011, Beirut, Lebanon
youssef.bassil@lacsc.org, mohammad.alwani@lacsc.org

## ABSTRACT

With the advent of digital optical scanners, a lot of paper-based books, textbooks, magazines, articles, and documents are being transformed into an electronic version that can be manipulated by a computer. For this purpose, OCR, short for Optical Character Recognition was developed to translate scanned graphical text into editable computer text. Unfortunately, OCR is still imperfect as it occasionally mis-recognizes letters and falsely identifies scanned text, leading to misspellings and linguistics errors in the OCR output text. This paper proposes a post-processing context-based error correction algorithm for detecting and correcting OCR non-word and real-word errors. The proposed algorithm is based on Google's online spelling suggestion which harnesses an internal database containing a huge collection of terms and word sequences gathered from all over the web, convenient to suggest possible replacements for words that have been misspelled during the OCR process. Experiments carried out revealed a significant improvement in OCR error correction rate. Future research can improve upon the proposed algorithm so much so that it can be parallelized and executed over multiprocessing platforms.

## 1. INTRODUCTION

The drastic introduction of modern computers into every area of life has radically led to a paradigm shift in the way people trade, communicate, learn, share knowledge, and get entertained. Present-day computers are electronic and digital, and thus they can only process data in digital format. Given that, anything that requires a computer processing must first be transformed into a digital form. For instance, the Boston Public Library which features more than 6.1 million books [1], all open to public, inevitably has to convert all its paper-based books into digital documents so that they can be stored on a computer's hard drive. In the same context, it has been estimated that more than 200 million books are being published every year [2], many of which are being distributed and printed on traditional papers [3]. In view of that, it is impossible to store all these books on a computer and manage them using software applications unless first converted into a digital form.

OCR, short for Optical Character Recognition is the process of converting scanned images of text into editable digital documents that can be processed, edited, searched, saved, and copied for an unlimited number of times without any degradation or loss of information using a computer. Although OCR sounds perfect for transforming a traditional library into an e-library, it is subject to errors and shortcomings. Practically, the error rate of OCR systems can fairly become high, occasionally close to 10% [4], if the papers being scanned have numerous defects such as bad physical condition, poor printing quality, discolored materials, and old age papers. When an OCR system fails to recognize a character, an OCR error is produced, commonly causing a spelling mistake in the output text. For instance, character "B" can be improperly converted into number "8", character "S" into number "5", character "O" into number "0", and so forth. To remedy this problem, humans can manually review and correct the OCR output text by hand. To a certain extent, this procedure is considered costly, time consuming, laborious, and error-prone as the human eye may miss some mistakes. A better approach, could be automating the correction of misspelled words using computer software such as spell checkers. This solution consists of using a lookup dictionary to search for misspelled words and correcting them suitably. While this technique tries to solve the actual problem, it in fact introduces another problem, yet more awkward. In effect, the dictionary approach tries to look at the misspelled word in isolation, in a sense that it does not take into consideration the context in which the error has occurred. For this reason, linguistic context-based error correction techniques were proposed to detect and correct OCR errors with respect to their grammatical and semantic context [5, 25]. As a result, the net outcome using context-based error correction can be noteworthy as it greatly improves the OCR error correction rate [6].

Obviously, all of the aforementioned methods have still a common drawback; they all require the integration of a vast dictionary of massive terms that covers almost every single word in the target language. Additionally, this dictionary should encompass proper nouns, names of countries and locations, scientific terminologies, and technical keywords. To end with, the content of this dictionary should be constantly updated so as to include new emerging words in the language. Since in practice it is almost impossible to compile such a wide-ranging dictionary, it would be wise using a web of online text corpuses containing all possible words, terms, expressions, jargons, and terminologies that have ever occurred in the language. This web of words can be seamlessly provided by Google search engine [30].

This paper proposes a new post-processing method for OCR error correction based on spelling suggestion and the "*did you mean*" feature of Google's online web search engine. The goal of this approach is to automate the proofreading of OCR text and provide context-based detection and correction of OCR errors. The process starts by chunking the OCR output text $B$, possibly containing spelling mistakes, into blocks of five words each. Then, every single block in $B = \{b_0, b_1, b_2...b_n\}$ is submitted as a search query to Google's web search engine; if the search returns "*did you mean: $c_i$*" where $c_i$ is the alternative spelling suggestion for block $b_i$, then block $b_i$ is considered misspelled and is replaced by the suggested block $c_i$. Otherwise, in case no suggestion is returned, block $b_i$ remains intact and is appended to the list of correct blocks. Eventually, the fully corrected OCR text is the collection of the correct blocks, formally represented as $C = \{ c_0, c_1, c_2...c_n \}$.

## 2. OPTICAL CHARACTER RECOGNITION

Optical Character Recognition (OCR) is the process of translating images of handwritten or typewritten text into machine-editable text [4]. These images are commonly captured using computer scanners or digital cameras. The quality of the images being scanned plays a critical role in determining the error rate in the recognized text. For instance, OCR systems may lead to poor and insignificant results if their input source is physically out of condition, of old age, having low printing quality, and containing imperfections and distortions such as rips, stains, blots, and discolorations [7, 8].

Two types of optical character recognition systems exist. The first type is the offline OCR system which extracts data from scanned images through optical scanners and cameras; while the second type is the online OCR system which employs special digitizers to capture in real-time the user's writing according to the order of the lettering, speed, and pen movements and strokes.

Technically speaking, every OCR system undergoes a process of sequential stages in order to convert a paper text document into a computer digital text. This process consists of the image acquisition stage which captures the input document; the pre-processing stage which improves the quality of and removes artifacts from the input document; the feature extraction and classification stage which extracts similar objects from the input document and groups them into classes so that they can be recognized as characters and words; and finally the post-processing stage which refines the OCR output text by correcting linguistic misspellings.

## 3. OCR POST-PROCESSING

As discussed in the previous section, post-processing is the last activity to occur in a series of OCR processing stages. Chiefly, the goal of post-processing is to detect and correct linguistic misspellings in the OCR output text after the input image has been scanned and completely processed. Fundamentally, there are two types of OCR errors: non-word errors and real-word errors [9]. A non-word error is a word that is recognized by the OCR system; however, it does not correspond to any entry in the lexicon. For instance, when "How is your day" is recognized by the OCR system as "Huw is your day", then "Huw" is said to be a non-word error because "Huw" is not defined in the English language. In contrast, a real-word error is a word that is recognized by the OCR system and does correspond to an entry in the lexicon, albeit it is grammatically incorrect with respect to the sentence in which it has occurred. For instance, when "How is your day" is recognized by the OCR system as "How is you day", then "you" is considered a real-word error because "you" although is syntactically correct (available in the English language), its usage in the sentence is grammatically incorrect. Typically, non-word and real-word errors fall under three classes of errors: deletion, insertion, and substitution errors. The deletion error occurs when one or more characters are discarded or removed from within the original word. For example, mis-recognizing the word "House" as "Hose", "Huse", "Hse", or even "ouse". The insertion error occurs when one or more extra characters are added or stiffed to the original word. For instance, mis-recognizing the word "Science" as "Sciencce" or even "Sciience". The substitution error occurs when one or more characters are accidently changed in the original word, such as changing the character "m" in "Computer" to "n" or changing the character "g" in "Against" to "q".

The poor condition of the papers being processed is by far the lone culprit for producing OCR errors and consequently causing OCR systems either to operate imprecisely or to fail utterly. Therefore, countless post-processing approaches and algorithms were proposed in an attempt to detect and correct OCR errors. In sum, they can be broadly broken down into three major categories: manual error correction, dictionary-based error correction, and context-based error correction.

### 3.1 Manual Error Correction

Intuitively, the easiest way to correct OCR errors is to hire a group of people to sit down and try to edit the OCR output text manually. This approach is often known as proofreading and although is straightforward, it requires a continuous manual human intervention. Distributed Proofreaders (DP) [10] initially initiated by Charles Franks in 2000 and originally meant to assist the Project Gutenberg (PG) [11], is a web-based project designed to facilitate the collaborative conversion and proofreading of paper books into e-books. The idea of DP is to employ volunteers from all around the world to compare scanned documents with their corresponding OCR texts. Proofreading and correction of OCR errors are done through several rounds by several people as necessary. Once the process is completed, the verified OCR texts are assembled together and added to the Project Gutenberg archive.

Despite the fact that proofreading is achievable, it is still considered error-prone as humans may unintentionally overlook or miss some mistakes. Furthermore, manual correction is to some degree regarded as a laborious, costly, and time-consuming practice.

### 3.2 Dictionary-Based Error Correction

In a relentless effort to find a way to better detect and correct misspelled words in OCR text, researchers conceived the dictionary-based error correction methodology, also known as lexical error correction. In this approach, a lexicon or a lookup dictionary is used to spell check OCR recognized words and correct them if they are misspelled [12]. In some cases, a list of candidates is generated to assist in the correction of misspelled words. For instance, the correction candidates for the error word "poposd", can be "opposed", "proposed", "pops", and "popes". In point of fact, several non-trivial dictionary-based error correction algorithms exist, one of which is the string matching algorithm that weights the words in a text using a distance metric representing various costs. The correction candidate with the lowest distance with respect to the misspelled word is the best to fit as a correction [13]. Another algorithm [14] demonstrated that using the language syntactic properties and the *n*-gram model can speed-up the process of generating correction candidates and ultimately picking up the best matching candidate. [15] proposed an OCR post error correction method based on pattern learning, wherein a list of correction candidates is first generated from a lexicon, then the most proper candidate is selected as a correction based on the vocabulary and grammar characteristics surrounding the error word. [16] proposed a statistical method for auto-correction of OCR errors; this approach uses a dictionary to generate a list of correction candidates based on the *n*-gram model. Then, all words in the OCR text are grouped into a frequency matrix that identifies the exiting sequence of characters and their count. The correction candidate having the highest count in the frequency matrix is then selected to substitute the error word. [17] proposed an improved design that employs a clustering technique to build a set of groups containing all correction candidates. Then, several iterations of word frequency analysis are executed on these clusters to eliminate the unlikely candidate words. In due course, only a single candidate will survive to replace the misspelled word. [18] proposed the use of a topic model to correct the OCR output text. It is a global word probability model, in which documents are labeled with a semantic topic having a specific independent vocabulary distribution. In other words, every scanned document is semantically classified according to its topic using unsupervised training model. Every misspelled word is then corrected by selecting the correction candidate that belongs to the same class of the actual error. [19] proposed a divergent approach based on syntactic and semantic correction of OCR errors; the idea pivots around the analysis of sentences to deduce whether or not they are syntactically and semantically correct. If a suspicious sentence is encountered, possible correction candidates are generated from a dictionary and grouped top-down with respect to their strongest syntactic and semantic constraints. In the long run, the candidate on the top of each group is the one that substitutes the corresponding OCR error. [20] proposed the idea of using a Hidden Markov Model (HMM) to integrate syntactic information into the post-processing error correction. The suggested model achieved a higher rate of error correction due to its statistical nature in selecting the most probable candidate for a particular misspelled word. [21] introduced an intelligent autonomic model able of self-learning, self-configuring, and self-adapting. The idea behind it is that as the system operates, as its ability to self-find and self-correct errors increases. [22] proposed a blend of post-processing tools that help fight against spelling errors. In this method, the OCR text is sent through a series of filters with the intention of correcting misspellings via multiple passes. On every pass, a spell checker tool intervenes to detect and correct misspelled words. After several passes, the number of OCR errors starts by exponentially getting reduced.

### 3.3 Context-Based Error Correction

Hypothetically, dictionary-based error correction techniques are reasonably plausible and successful. However, they are unable to correct errors based on their context, i.e. correcting errors based on their grammatical occurrence in the sentence. Context-based error correction techniques, on the other hand, perform error detection and correction based on the error grammatical and sometimes semantic context. This would solve the previous dilemma of correcting real-word errors such as in the sentence "How is you day", because according to the context in which "you" has occurred, it is unlikely to have a personal pronoun followed by a noun, rather, it is more likely to have a possessive pronoun followed by a noun.

In order to bring context-based error correction into practice, several innovative solutions were considered, the majority of them are grounded on statistical language models (SLMs) and feature-based methods. [23] described a context-sensitive word-error correction system based on confusion mapping that uses confusion probabilities to identify frequently wrong sequences and convert them into the most probable correct sequence. In other terms, it models how likely one letter has been misinterpreted as another. [24] applied a part-of-speech (POS) tagger and the grammatical rules of the English language to capture real-word errors in the OCR text. For instance, one of these rules states that a verb can be followed by a gerund object but it cannot be followed by a second verb, while another rule states that a third person verb in the present tense must always take an "s". The aggregate of these rules drove the logic of the algorithm and achieved a reasonable context-based OCR error correction. [25] used word trigrams to capture and correct non-word and real-word errors. The idea is to use a combination of a lookup dictionary to correct non-word errors, and a statistical model to correct real-word errors according to their context. [26] proposed a Bayesian classifier that treats the real-word errors as ambiguous, and then tries to find the actual target word by calculating the most likely candidate based on probabilistic relationships between the error and the candidate word. [27] joined all the previous ideas into a concrete solution; it is a POS tagger enhanced by a word trigram model and a statistical Bayesian classifier developed to correct real-word errors in OCR text. Overall, the mixture of these techniques hugely improved the OCR post-processing error correction rate.

## 4. LIMITATIONS OF DICTIONARY-BASED ERROR CORRECTION

Although dictionary-based error correction techniques are easy to implement and use, they still have various limitations and drawbacks that prevent them from being the perfect solution for OCR post-processing error correction.

The first limitation is that dictionary-based approach requires a wide-ranging dictionary that covers every single word in the language. For instance, the Oxford dictionary [28] embraces 171,476 words in current use, and 47,156 obsolete words, in addition to their derivatives which count around 9,500 words. This suggests that there is, at the very least, a quarter of a million distinct English words. Besides, spoken languages may have one or more varieties each with dissimilar words, for instance, the German language has two varieties, a new-spelling variance and an old-spelling variance. Likewise, the Armenian language has three varieties each with a number of deviating words: Eastern Armenian, Western Armenian, and Grabar. The Arabic language also follows the same norm as it has many assortments and dialects that diverge broadly from country to country, from region to region, and from era to era [29]. For instance, the ancient Arabic language that was used before 600 A.D. in the north and south regions of the Arabian Peninsula is totally different from the classical Arabic that is being used in the present-day. Therefore, it is obvious that languages are not uniform, in a sense that they are not standardized and thereby cannot be supported by a single dictionary.

The second limitation is that regular dictionaries normally target a single specific language and thus they cannot support multiple languages simultaneously. For instance, the Oxford and the Merriam–Webster dictionaries only target the English language. The Larousse dictionary targets the French language, while the Al Munjid dictionary targets the Arabic language. Henceforth, it is unquestionably impossible to create an international dictionary pertaining to all languages of the world.

The third limitation is that conventional dictionaries do not support proper and personal names, names of countries, regions, geographical locations and historical sites, technical keywords, domain specific terms, and acronyms. For instance, an ordinary dictionary could falsely detect "Thomas Jefferson", "Machu Picchu", and "São Tomé" as incorrect words. Similarly, scientific terminologies such as "RAM", "CPU", and "pixel", and names of diseases such as "AIDS", "Hypothermia", and 'Malaria' could incorrectly be detected as misspellings. In total, it is nearly unviable to compile a universal dictionary with words from all existing domains and disciplines.

The fourth and last limitation is that the content of a standard dictionary is static in a way that it is not constantly updated with new emerging words unless manually edited, and thus, it cannot keep pace with the immense dynamic breeding of new words and terms.

For all the above reasons, attaining better OCR post-processing dictionary-based error correction fallouts, greatly require finding a universal, multi-language, and dynamic dictionary embracing a colossal volume of entries, words, terms, proper nouns, expressions, jargons, and terminologies that possibly could occur in a text.

## 5. PROPOSED SOLUTION

This paper proposes a new post-processing method and algorithm for OCR error correction based on the "*did you mean*" spelling suggestion feature of Google's online web search engine [30]. The idea centers on using Google's massive indexed data to detect and correct misspelled words in the OCR output text. The algorithm starts first by chopping the OCR text into several tokens of words. Then, each token is sent as a search query to Google's search engine so that it gets processed. In case the query contains a misspelled word, Google will suggest a possible correction via its "*did you mean*" feature. Consequently, this spelling suggestion is to be considered as a correction for the misspelled query.

### 5.1 Inner Workings of Google's Spelling Suggestion

According to [31, 32], Google's spelling suggestion system can suggest an alternative correct spelling for the often made typos, misspellings, and keyboarding errors. Under the hood, Google has a titanic database of millions of public web pages containing trillions of term collections and *n*-gram words that can be used as groundwork for all kinds of linguistic applications such as machine translation, speech recognition, and spell checking, as well as other types of text processing problems. Inherently, Google's spelling suggestion scheme is based on the probabilistic *n*-gram model originally proposed by [33] for predicting the next word in a particular sequence of words. In short, an *n*-gram is simply a collocation of words that is *n* words long, for instance, "The boy" is a 2-gram phrase also referred to as bigram, "The boy scout" is a 3-gram phrase also referred to as trigram, "The boy is driving his car" is a 6-gram phrase, and so forth. Google's algorithm automatically examines every single word in the search query for any possible misspelling. It tries first to match the query, basically composed of ordered association of words, with any occurrence alike in Google's index database. If the number of occurrence is high, then the query is considered correct and no correction is to take place. However, if the query was not found, Google tries to infer the next possible correct word in the query based on its *n*-gram statistics deduced from Google's database of indexed webpages. In due course, an entire suggestion for the whole misspelled query is generated and displayed to the user in the form of "*did you mean: spelling-suggestion*". For example, searching for the word "conputer" drives Google's search engine to suggest "*did you mean: computer*". Likewise, searching for "conputer on the tesk" drives Google's search engine to suggest "*did you mean: computer on the desk*". Also trying to search for the proper name "maw tsi toung" drives Google's search engine to suggest "*did you mean: mao tse tung*". Figure 1-3 show the different suggestions returned by Google's search engine when searching for the misspelled queries "conputer", "conputer on the tesk", and "maw tsi toung" respectively.

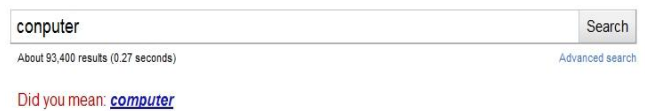

**Figure 1:** Spelling suggestion for "conputer"

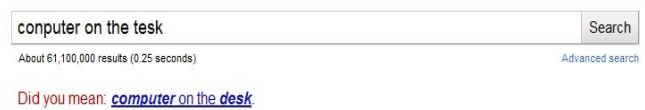

**Figure 2:** Spelling suggestion for "conputer on the tesk"

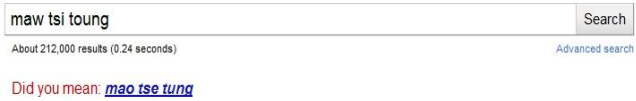

**Figure 3:** Spelling suggestion for "maw tsi toung"

## 5.2 Proposed Solution Specifications

The proposed solution is a context-based error correction algorithm built on Google's spelling suggestion technology, and meant to be integrated into OCR systems as a post-processor to detect and correct non-word and real-word errors. Specifically, and after the image document has been scanned and digitally processed to produce computer text, the proposed algorithm breaks down the OCR recognized output text into a collection of blocks, each containing five words (Five words were chosen just to provide Google with enough insights about the context of every block). These blocks are fed one by one to Google's search engine as search parameters. If Google returns a successful response without the "*did you mean*" expression, then it is evident that the query contains no misspelled words and thus no correction is needed for this particular block of words. Contrariwise, if Google responds with a "*did you mean: spelling-suggestion*" expression, then definitely the query contains some misspelled words and thus a correction is required for this particular block of words. The actual correction consists of replacing the original block in the OCR output text by the Google's alternative suggested correction. Figure 4 is a variation of the generic OCR system proposed by [4], however, upgraded with an additional post-processing layer using the proposed error correction algorithm.

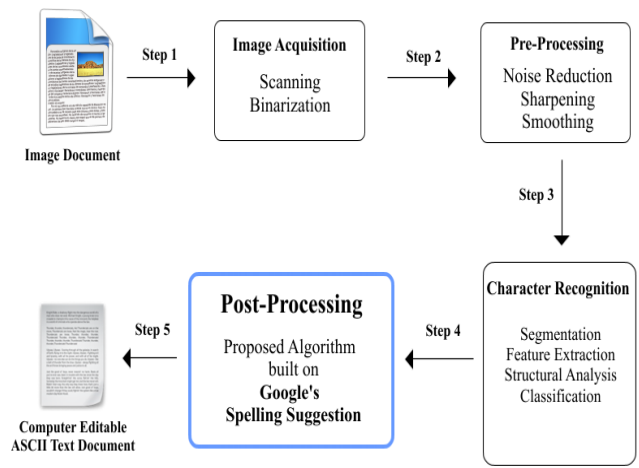

**Figure 4:** OCR system enhanced with the proposed algorithm

## 5.3 The Proposed Algorithm

The proposed algorithm comprises several steps to be executed in order to correct OCR misspellings. The algorithm takeoffs by dividing the OCR output text $B$ into a series of blocks $b_0, b_1, b_2...b_n$ each made out of five words. Subsequently, blocks in $B=\{ b_0, b_1, b_2...b_n \}$ are sent sequentially, one by one to Google's online web search engine as a search query parameters. The search results returned by Google are then parsed to identify whether or not they contain a "*did you mean: $c_i$*" expression, where $c_i$ is the spelling suggestion for block $b_i$. If true, then the block $b_i$ must contain a misspelled word, and hence, $c_i$ is extracted from "*did you mean: $c_i$*" and appended to a text file $C$ which will eventually hold the entire correct blocks $C=\{ c_0, c_1, c_2...c_n \}$. In contrast, if the search results did not contain any "*did you mean*" expression, then the block $b_i$ is said to contain no misspelled words, and thus, the original $b_i$ is added intact to the text file $C$. Ultimately, when all blocks get validated, the OCR post-processing stage finishes and the algorithm halts. The text file $C$ holding the complete corrected OCR text can now be safely handled appropriately, i.e. printed, saved, or edited using a word processor. Figure 5 is a flowchart summarizing the various computational steps of the proposed algorithm, executed to detect and correct misspellings in OCR text.

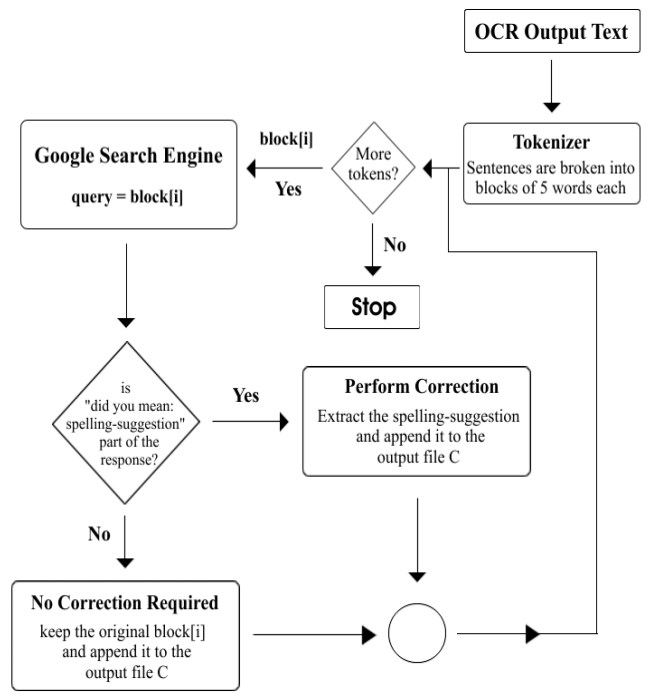

**Figure 5:** The executional steps of the proposed algorithm

## 5.4 The Pseudo-Code

The following pseudo-code is a high-level computing platform-independent description of the entire logic behind the proposed algorithm.

```
// the purpose of this function is to correct the spelling errors in the
   OCR output text using the Google's spelling suggestion feature.
// INPUT: OCR plain text received from the previous character
   recognition stage.
// OUTPUT: Corrected OCR text containing the least possible
   misspellings.

BEGIN

Function Post-Correction (ocr_text)
{
    // breaks the ocr_text into block of 5 words each
    B ← Tokenize(ocr_text , 5)

    for (i←0 to N)  // iterates until all tokens are exhausted
    {
        // sends every B[i] to Google search engine
        result ← GoogleSearch(B[i])

        if(result contains "did you mean")
        {
```

```
           // indicates some misspellings in B[i]
           // parses the search results, extracts the suggestion
           C ← ExtractCorrection(result)
       }
    else   // and appends it to the output file C
           C ← B[i]
           // no misspellings were found so add the original B[i] to C
    }
    RETURN C  // file C now holds the complete corrected OCR text
}
END
```

The function *Post-Correction()* contains one for loop that is executed *n* times, where *n* is the total number of tokens in the OCR text. Considering "*result ← GoogleSearch(B[i])*" as the basic operation, the time complexity of the algorithm is described as follows:

$$\sum_{i=0}^{n} 1 = n$$ and thus the algorithm is of time complexity $O(n)$

Since the basic operation is to be executed *n* times irrespectively of the content of the input OCR text, $C_{Best}(n) = C_{Worst}(n) = C_{Average}(n) = n$

## 6. EXPERIMENTS & RESULTS

In the experiments, OCR was performed on two low-quality image documents, each in a different language: English [34] and Arabic [35]. *Google.com* was used to spell-check the English document, while *Google.ae* was used to spell-check the Arabic document. Additionally, one of the most renowned proprietary OCR software solutions, the *OmniPage version 17* by Nuance Communications [36] with English and Arabic support was utilized to carry out the OCR process. The proposed algorithm was implemented using MS C# 4.0 under the MS .NET Framework 4.0 and the MS Visual Studio 2010.

Figure 6 shows the original English document to be processed. The subsequent Table I delineates all misspellings (underlined) generated during the OCR process. Next is Table II, which outlines the same OCR text of Table I, however, error-corrected using the proposed OCR post-processing error correction algorithm.

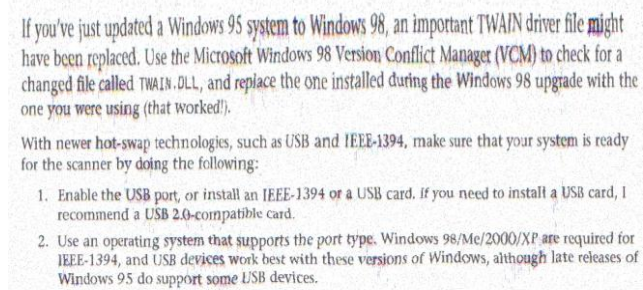

**Figure 6:** Low-quality English image document

**Table I:**
The results of performing OCR on the English document

| |
|---|
| If yotrve gust updated a Windows 95 symtem to Window 98, an important TWAIN driver file night have been replaced. Use the MicroSeoft windows 98 Verslon confliqt Mmage ,(VCM) to gheck fmr a changed file celled TWAIN.DLL, and repiace the one installed during Windows 98 upgrade with the one yu were using (that worked!). With newer hut-swap technralngies, such as USB and lEEE -1394, make sure that your sysilem is ready for the scanner by aoing the following; 1. Enable the USB port, or install an [EEE-1394 or a USB card. if you need to install a USB card, I reccommend a USB 2.o - compatihle card. 2. Use an operating system that supports the port type. Windows 95 Me 2000 XP are required for IEIiE-1394 and USB devices wmrk best with these versions of Windows, although late releases of Windows 95 do support some ESB devices. |

**Table II:**
The results obtained after applying the proposed error correction algorithm

| |
|---|
| If you've just updated a Windows 95 system to Window 98, an important Twain driver file might have been replaced. Use the Microsoft windows 98 Version conflict Mage ,(VCM) to check for a changed file called TWAIN.DLL, and replace the one installed during Windows 98 upgrade with the one you were using (that worked!). With newer hut-swap technologies, such as USB and IEEE -1394, make sure that your system is ready for the scanner by using the following; 1. Enable the USB port, or install an IEEE-1394 or a USB card. if you need to install a USB card, I recommend a USB 2.o - compatible card. 2. Use an operating system that supports the port type. Windows 95 Me 2000 XP are required for IEEE 1394, and USB devices work best with these versions of Windows, although late releases of Windows 95 do support some USB devices. |

The OCR text delineated in Table I comprehended 27 misspelled words out of 126 (the total number of words in the whole original text), making the error rate close to E = 27/126 = 0.214 = 21.4%. Several of these errors were proper names such as "Microsoft" and "IEEE", while others were technical words such as "USB" and "TWAIN". The remaining errors were regular English words such as "recommend", "compatible", "work", "might", etc. Table II exposed the results of post-processing the OCR output text in Table I using the proposed error correction algorithm. 23 misspelled words out of 27 were corrected, leaving only 4 non-corrected errors, and they are "Mmage" which was falsely corrected as "Mage", "agoing" as "using", whereas "Hut" and "2.o" were not corrected at all. As a result, the error rate using the proposed algorithm dropped to E = 4/126 = 0.031 = 3.1%. Consequently, the improvement can be calculated as I = 0.214/0.031 = 6.90 = 690%, that is increasing the error correction rate by a factor of 6.9.

The following is the Arabic document to be processed and tested. Figure 7 depicts the original document, while Table III delineates the OCR results along with the numerous misspelled words that were generated. Table IV shows the results of post-processing the OCR output text of Table III using the proposed error correction algorithm.

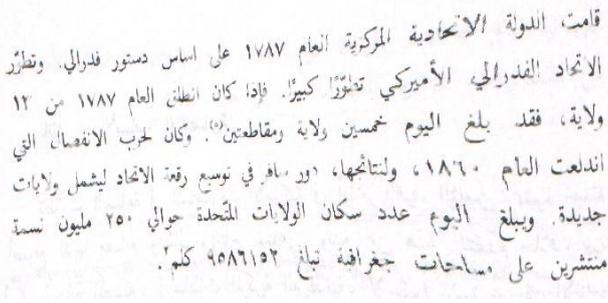

**Figure 7:** Low-quality Arabic image document

**Table III:**
The results of performing OCR on the Arabic document

| |
|---|
| قامت اندولة الاتحادية المركرية العام ١٧٨٧ على اساس دستور فدرالي. وتطوّر الاتحاد الفدرالي الأميركي تضورا كبيرا. فإذا كان نصالق العام ١٧٨٧ من ١٣ ولاية، فقد بلغ اليوم خمسين ولاية ومقاطعتين. وكان لحرب الانفصال التي اندلعت العام ١٨٦٠، نور سافر في توسيع رقعى الاتحاد ليشمل ولايات جديدة. ويبلغ اليوعم عدد سكان الولايات المتّحدة حوالي ٢٥٠ مليون نسمة منتشرين على مساحات جغرافية تبلغ ٩٥٨٦١٥٢ كلم |

**Table IV:**
The results obtained after applying the proposed error correction algorithm

| |
|---|
| قامت الدولة الاتحادية المركزية العام ١٧٨٧ على اساس دستور فدرالي. وتطوّر الاتحاد الفدرالي الأميركي تطورا كبيرا. فإذا كان نصف العام ١٧٨٧ من ١٣ ولاية، فقد بلغ اليوم خمسين ولاية ومقاطعتين. وكان لحرب الانفصال التي اندلعت العام ١٨٦٠، نور سافر في توسيع رقعة الاتحاد ليشمل ولايات جديدة ويبلغ اليوم عدد سكان الولايات المتّحدة حوالي ٢٥٠ مليون نسمة منتشرين على مساحات جغرافية تبلغ ٩٥٨٦١٥٢ كلم |

The OCR text delineated in Table III comprehended 8 misspelled words out of 64 (the total number of words in the whole original text), making the error rate close to E = 8/64 = 0.125 = 12.5%. The majority of these errors were regular Arabic words such as "مساحات", "المركزية", "الدولة", etc. Table IV exposed the results of post-processing the OCR output text in Table III using the proposed error correction algorithm. 6 misspelled words out of 8 were corrected, leaving only 2 non-corrected errors, and they are "نصالق" which was falsely corrected as "نصف", and "نور" was not corrected at all. As a result, the error rate using the proposed algorithm dropped to E = 2/64 = 0.031 = 3.1%. Consequently, the improvement can be calculated as I = 0.125/0.031 = 4.03 = 403%, that is increasing the error correction rate by a factor of 4.

## 7. EXPERIMENTS EVALUATION

The experiments conducted on the proposed OCR post-processing error correction algorithm clearly revealed an error detection and correction improvement of 690% for English text and 403% for Arabic text. In other words, 6.9 times more English errors and 4 times more Arabic errors were detected and corrected. On average, the proposed algorithm improved the error correction rate by I= (609% + 403%) / 2 = 506%, that is increasing the overall error correction rate by a factor of 5.06. Table V roughly sketches the head-to-head OCR experimental results between the proposed algorithm and the *OmniPage* software suite.

**Table V:**
Head-to-head comparison between the proposed algorithm and the OmniPage suite

| | English Document Total words = 126 | Arabic Document Total words = 64 |
|---|---|---|
| Number of errors resulted from using OmniPage 17 | 27 | 8 |
| Number of errors resulted from using the proposed algorithm | 4 | 2 |
| OmniPage error rate | 21.4% | 12.5% |
| Proposed algorithm error rate | 3.1% | 3.1% |
| Proposed algorithm improvement ratio | 6.9 (690%) | 4.03 (403%) |

## 8. CONCLUSIONS

This paper presented a new post-processing technique for OCR error detection and correction based on Google's online spelling suggestion. Since Google is a giant warehouse of indexed real-world pages, articles, blogs, forums, and other sources of text, it can suggest common spellings for words not found in standard dictionaries. The proposed algorithm exploited Google's *"did you mean"* technology with the purpose of using query spelling suggestions to correct non-word and real-word errors in OCR output text. Experiments undertaken showed a sharp improvement in OCR error correction rate as higher number of misspellings and linguistic errors were detected and corrected using the proposed method compared with other traditional existing ones.

## 9. FUTURE WORK

As further research, the proposed algorithm can be re-designed to support multiprocessing platforms so as to operate in a parallel fashion over a bunch of concurrent processors or even over a bunch of distributed computing machines. The expected results would be a faster algorithm of time complexity $O(n/m)$, where $n$ is the total number of word tokens to be spell-checked and $m$ is the total number of processors.

## ACKNOWLEDGMENTS

This research was funded by the Lebanese Association for Computational Sciences (LACSC), Beirut, Lebanon under the "Web-Scale OCR Research Project – WSORP2011".